\documentclass[letterpaper, 10pt, conference]{ieeeconf}

\IEEEoverridecommandlockouts                              

\usepackage[noadjust]{cite}
\usepackage{grffile}
\usepackage{multicol}
\usepackage[fleqn]{amsmath}
\usepackage{amssymb}
\usepackage{comment}
\usepackage{color}
\usepackage{graphicx}
\usepackage{mathtools}
\usepackage[ruled,linesnumbered]{algorithm2e}
\usepackage{float}
\usepackage{setspace}
\usepackage{breqn}
\usepackage{makecell}
\usepackage{multirow}
\usepackage{booktabs}
\usepackage{afterpage}
\usepackage{subcaption}
\usepackage{siunitx}
\usepackage{import}
\makeatletter
\let\NAT@parse\undefined
\makeatother
\usepackage[hidelinks,bookmarks=true]{hyperref}
\hypersetup{final}

\allowdisplaybreaks

\newcolumntype{M}[1]{>{\centering\arraybackslash}m{#1}}

\DeclareMathOperator{\RMSE}{RMSE}

\newcommand{\argmin}{\operatornamewithlimits{argmin}}
\newcommand{\argmax}{\operatornamewithlimits{argmax}}

\def\argmax{\mathop{\rm argmax}}
\def\argmin{\mathop{\rm argmin}}

\def\atan2{\mathop{\rm atan2}}

\title{\LARGE \bf
A Maximum Likelihood Approach to Extract Finite Planes \\
from 3\nobreakdash-D Laser Scans}

\author{Alexander~Schaefer, Johan~Vertens, Daniel~B{\"u}scher, Wolfram~Burgard
\thanks{\copyright\ 2018 IEEE. Personal use of this material is permitted.  Permission from IEEE must be obtained for all other uses, in any current or future media, including reprinting/republishing this material for advertising or promotional purposes, creating new collective works, for resale or redistribution to servers or lists, or reuse of any copyrighted component of this work in other works.}
\thanks{This work has been partially supported by Samsung Electronics~Co.~Ltd. under the GRO program.}%
\thanks{All authors are with the Department of Computer Science, University of Freiburg, Germany.}%
\thanks{\tt \small \{aschaef, vertensj, buescher, burgard\} @cs.uni-freiburg.de}}%

\begin{document}

\maketitle
\thispagestyle{empty}
\pagestyle{empty}

\begin{abstract}
Whether it is object detection, model reconstruction, laser odometry, or point cloud registration: 
Plane extraction is a vital component of many robotic systems.
In this paper, we propose a strictly probabilistic method to detect finite planes in organized 3\nobreakdash-D laser range scans.
An agglomerative hierarchical clustering technique, our algorithm builds planes from bottom up, always extending a plane by the point that decreases the measurement likelihood of the scan the least.
In contrast to most related methods, which rely on heuristics like orthogonal point-to-plane distance, we leverage the ray path information to compute the measurement likelihood.
We evaluate our approach not only on the popular SegComp benchmark, but also provide a challenging synthetic dataset that overcomes SegComp's deficiencies.
Both our implementation and the suggested dataset are available at~\cite{ppe2019}.
\end{abstract}

\section{Introduction}
\label{sec:introduction}

The geometry of many man-made environments like factory floors, offices, and households can be described by a set of finite planes.
Robots navigating these types of environments often rely on 3\nobreakdash-D laser range finders, which capture up to millions of reflections per second.
Plane extraction methods take these highly redundant raw sensor measurements and reduce them to the parameters of the underlying planes, thus reducing the computational effort and the memory footprint required for processing the sensor data.
Plane extraction may also increase accuracy in tasks like scan matching and sensor calibration, and it enables applications like model reconstruction and object detection in the first place.

\begin{figure}
	\centering
	\begin{subfigure}[t]{0.49\linewidth}
		\includegraphics[width=\linewidth]{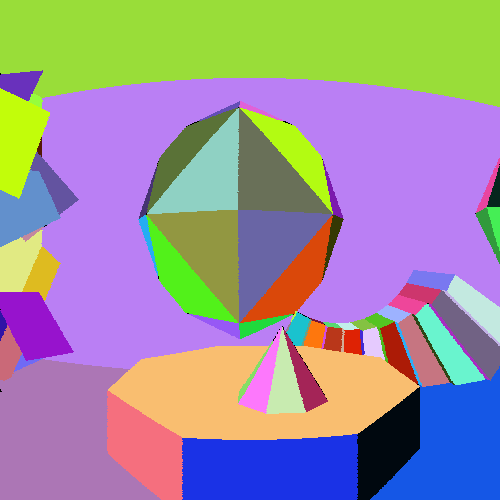}
		\caption{Ground-truth segmentation.}
		\label{fig:synpeb_seg_gt}
	\end{subfigure}
	\hfill
	\begin{subfigure}[t]{0.49\linewidth}
		\includegraphics[width=\linewidth]{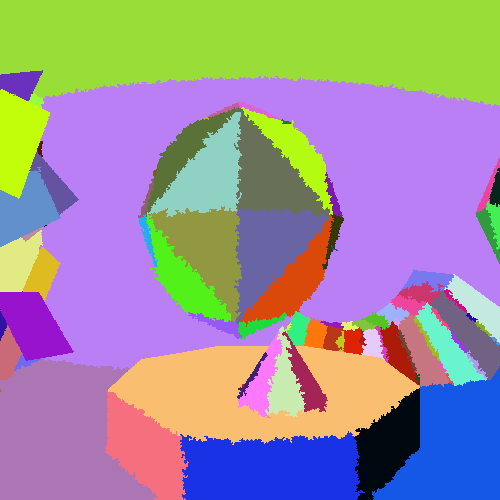}
		\caption{PPE segmentation.}
		\label{fig:synpeb_seg_ppe}
	\end{subfigure} 
	\caption{Ground-truth segmentation of an organized  \mbox{$500 \times 500$} point cloud taken from the suggested SynPEB dataset and segmentation result of PPE, the proposed method.}
	\label{fig:synpeb}
\end{figure}

The presented method, dubbed probabilistic plane extraction~(PPE), extends our recent work on polyline extraction from 2\nobreakdash-D laser range scans~\cite{schaefer2018} to three dimensions.
Essentially, PPE is a maximum likelihood approach based on agglomerative hierarchical clustering.
In the beginning, PPE represents the scan by a large set of planes -- one plane for every reflection -- and then iteratively merges them, in each step choosing the subset whose merger maximizes the measurement likelihood of the whole scan, until a specified stopping criterion is met.
Figure~\ref{fig:synpeb_seg_ppe} shows an exemplary segmentation result.

Our approach distinguishes itself from the large body of related work in two respects.
First, all methods surveyed in the following resort to heuristics like orthogonal distance between ray endpoint and plane when estimating the measurement likelihood of a scan conditioned on a set of planes.
Instead, PPE accounts for the true ray path from start to end.
This more accurate sensor model leads to more accurate results, as demonstrated by our experiments.
Second, due to its probabilistic formulation, PPE requires only one robust parameter to control the granularity of the extracted planes.
In contrast, some of the surveyed methods need up to a dozen carefully tuned parameters in order to obtain reasonable results.

\section{Related Work}
\label{sec:related_work}

This section provides an overview over the state of the art considering plane extraction from 3\nobreakdash-D lidar scans.
We distinguish four classes of approaches: region growing, clustering, random sample consensus~(RANSAC), and the Hough transform.

In a nutshell, region growing first selects some seed points from the input point cloud, which are then grown into regions by iteratively adding all neighboring points that pass a set of criteria.
Hoover et~al.~\cite{hoover1996}, for instance, select the points with the highest local planarity score as seeds.
During the growing process, they add all adjacent points to the regions that do not exceed a specified maximum difference of normals, Euclidean distance, and orthogonal distance.
In contrast, H{\"a}hnel et~al.~\cite{haehnel2003} choose seeds at random and grow planar polygons by including all neighboring points that do not push the mean squared error of the resulting plane over a given limit.

Deschaud et~al.~\cite{deschaud2010} propose an adaption of region growing to large noisy datasets.
They compensate for noise by introducing a filter that improves the estimation of endpoint normals, select seeds based on local planarity, and employ a voxel-based variant of region growing.
Nurunnabi et~al.~\cite{nurunnabi2012}, in turn, address noise by computing endpoint features like normals and curvature via a robust variant of principal component analysis~(PCA).
In another take on plane extraction from noisy point clouds, Dong et~al.~\cite{dong2018} combine region growing with energy optimization, where the energy is defined as the sum of geometric errors, spatial coherence, and the total number of planes.

Holz et~al.~\cite{holz2013} focus on plane extraction for time-sensitive applications.
Their method computes normal and curvature estimates not directly based on the point cloud, but based on an approximate mesh.
CAPE, an algorithm developed by Proen{\c{c}}a et~al.~\cite{proenca2018}, achieves even higher plane extraction rates at the expense of reduced accuracy.
First, the algorithm creates a low-resolution grid, pools the points in each cell, and applies PCA to each cell.
CAPE then grows regions composed of cells based on their PCA features.

Inspired by the observation that every line-shaped sequence of points in a laser scan is caused by a planar surface, Jiang et~al.~\cite{jiang1994}, Hoover et~al.~\cite{hoover1996}, and Cabo~et~al.~\cite{cabo2015} apply region growing to line segments instead of points.

As opposed to region growing, clustering extracts planes without the need to find suitable seed points.
Trevor et~al.~\cite{trevor2013}, for example, assign the same label to adjacent points of an organized range scan if the difference of their normals and their orthogonal distance falls below a given threshold, and subsequently extract planes by clustering points with the same labels.
Feng et~al.~\cite{feng2014} present a clustering algorithm that extracts planes from an organized point cloud with minimal latency.
It divides the point cloud uniformly into rectangular point groups, discards all non-planar groups, and subjects the remaining groups to agglomerative hierarchical clustering, using the mean squared orthogonal point-to-plane fitting error as clustering metric.
Eventually, it refines the extracted coarse planes by region growing.
Marriott et~al.~\cite{marriott2017} also cluster groups of coplanar points based on mean squared error, but instead of using a regular grid to define initial point groups, they propose an expectation-minimization algorithm that fits a Gaussian mixture model to the points.

Pham et~al.~\cite{pham2016} combine clustering and region growing.
They use region growing to oversegment the point cloud and then merge the resulting plane hypotheses via clustering, in each step minimizing an energy function that favors mutually parallel or orthogonal plane pairs.

RANSAC, initially developed by Fischler et~al.~\cite{fischler1981}, is a versatile iterative model fitting algorithm.
When applied to plane extraction, it selects three laser endpoints at random, fits a plane to them, searches for all points within a certain orthogonal distance, and determines the plane's fitness based on the corresponding point-to-plane distances.
This process is repeated until the algorithm finds a plane that satisfies a given minimal fitness.
Several works improve on standard RANSAC to overcome its deficiencies.
The robust estimator formulated by Gotardo et~al.~\cite{gotardo2003} counteracts RANSAC's tendency to disregard small regions.
Gallo et~al.~\cite{gallo2011} address the problem of RANSAC often connecting nearby patches that are actually unconnected, for example at steps and curbs.
By combining RANSAC with conformal geometric algebra, Sveier et~al.~\cite{sveier2017} perform the least squares fitting necessary to assess the fitness of a plane hypothesis analytically instead of numerically.
Alehdaghi et~al.~\cite{alehdaghi2015} present a highly parallelized GPU implementation of RANSAC for plane extraction.

Another general model fitting method, the Hough transform computes for each point in the discretized space of model parameters the fitness of the associated model instance given the data.
Vosselman et~al.~\cite{vosselman2004} describe how to apply this method to the problem of plane extraction from 3\nobreakdash-D point clouds.
Oehler et~al.~\cite{oehler2011} present a multi-resolution approach based on both the Hough transform and RANSAC.
For a review of further flavors of Hough transform-based plane extraction, the reader is referred to the review composed by Borrmann et~al.~\cite{borrmann2011}.

\section{Approach}
\label{sec:approach}

In this work, we present probabilistic plane extraction~(PPE), an approach to extract finite planes from organized 3\nobreakdash-D lidar scans.
PPE is a maximum likelihood estimation technique based on agglomerative hierarchical clustering.
As a maximum likelihood estimation technique, it searches for the set of planes that maximize the measurement probability of the given laser scan.
As an agglomerative clustering method, it attempts to find this set by creating a plane for each reflection first.
This plane explains the corresponding reflection perfectly.
PPE then reduces the number of planes by iteratively merging the set of adjacent planes whose merger maintains the highest measurement likelihood of the scan.
Clustering ends as soon as a given stopping criterion is met.

In the following, we first introduce the probabilistic sensor model, on the basis of which we then formulate plane extraction as a maximum likelihood estimation problem.
We describe in detail how our agglomerative hierarchical clustering algorithm solves this optimization problem, and finally explain the pseudocode.

\subsection{Probabilistic Sensor Model}

The sensor model tells the measurement probability of a 3\nobreakdash-D lidar scan given a set of planes. 
We denote the scan \mbox{$Z \coloneqq \{z_k\}$}, where \mbox{$k \in \{1,2,\ldots,K\}$} represents the index of a laser ray.
A single laser measurement~\mbox{$z \coloneqq \{s,v,r\}$} is composed of two three-element Cartesian vectors and a scalar: the starting point~$s$ of the ray, the normalized direction vector~$v$, and the ray length~$r$.
The set of finite planes~$L \coloneqq \{l_j\}$ extracted from the scan consists of a total of $J$ elements.
Each plane is represented by a three-element Cartesian support vector~$x$, a three-element Cartesian normal vector~$n$, and a set~$Q$ of ray indices: \mbox{$l \coloneqq \{x,n,Q\}$}.
While $x$ and $n$ define the location and orientation of the plane, $Q$ determines its extent.
This representation can not only handle convex planes, but also concave planes or planes with holes.

Most lidar sensors exhibit approximately normally distributed noise in radial direction and relatively small angular noise.
Consequently, we neglect angular noise and model the distribution of the measured length of a single ray conditioned on a set of planes as a Gaussian probability density function centered at the true ray length:
\begin{align}
	p(z \mid L) = \mathcal{N}(r; \hat{r}(s,v,L), \sigma^2). 
	\label{eq:pzL}
\end{align}
Here, the function $\hat{r}(s,v,L) \in \mathbb{R}^+$ computes the distance between the starting point of the ray and the first intersection of its axis and all planes in~$L$.
The standard deviation~$\sigma$ of the radial noise is a function of multiple parameters such as sensor device, reflecting surface, and temperature, but usually not range.

By assuming independence between the individual laser rays, we can derive the measurement probability of the whole scan from equation~\eqref{eq:pzL} as
\begin{align*}
	p(Z \mid L) = \prod_{k=1}^K p(z_k \mid L).
\end{align*}

To our knowledge, we are the first to apply the above sensor model to plane extraction.
Most surveyed works model the measurement probability of a ray as a zero-centered normal distribution over the shortest distance between the measured ray endpoint and the nearest plane.
This heuristic does not account for the ray path, which leads to two undesired effects.
First, the nearest plane is not always the one that intersects the ray.
Second, the accuracy of the computed distance strongly depends on the incidence angle of the ray.

\subsection{Maximum Likelihood Estimation}
\label{sec:plane_extraction}

With the above sensor model, we formulate plane extraction as the following maximum likelihood estimation problem:
Find the set of planes~$L^*$ that maximizes the measurement probability of the whole scan~\mbox{$p(Z \mid L)$}.
The solution is trivial:
For each reflection in the laser scan, create a tiny plane that is not parallel to the ray and that intersects the ray at the measured ray length~$r$.
This solution, however, is merely a different representation of the raw lidar data.
In order to extract meaningful planes from the scan, we need to reduce the number of planes by constraining the optimization problem.
For the following derivation, we choose the maximum number of planes $J_{\max}$ as constraint parameter.
Note, however, that our approach allows us just as well to use arbitrary metrics like the maximum mean squared error of the ray radii or the Akaike Information Criterion~\cite{akaike1998}.
Formally, we are confronted with the constrained least squares optimization problem
\begin{align}
	L^* 
		&= \argmax_L p(Z \mid L)\Big|_{J(L) \leq J_{\max}} \notag \\
		&= \argmin_L -\log\Big( p(Z \mid L) \Big)\Big|_{J(L) \leq J_{\max}} 
			\label{eq:Lstar} \\
		&= \argmin_L \sum_{k=1}^K 
			\Big( r_k - \hat{r}(s_k,v_k,L) \Big)^2\Big|_{J(L) \leq J_{\max}} 
			\notag \\
		&\eqqcolon \argmin_L E(Z,L)\Big|_{J(L) \leq J_{\max}} \notag
\end{align}
where $J(L)$ is a function that determines the number of planes in $L$.
The transition from the second to the third line implies our assumption that all rays exhibit the same radial noise.
Hereafter, we will refer to~$E$ simply as the error of the set of planes~$L$.

Solving \eqref{eq:Lstar} is primarily a combinatorial problem.
Even if we knew the parameters $\{x_j\}$ and $\{n_j\}$ of the planes, we would still not know the data associations~$\{Q_j\}$, i.e. which rays belong to which plane.
Exhaustively searching the space of all data associations for the combination that maximizes the measurement probability quickly leads to combinatorial explosion even for small~$J_{\max}$.
PPE solves this problem via agglomerative hierarchical clustering.

\subsection{Agglomerative Hierarchical Clustering}

In its generic form, agglomerative hierarchical clustering builds clusters from bottom up:
The algorithm first assigns each observation its own cluster and then iteratively merges adjacent pairs of clusters.
In each iteration, it decides which pair to merge based on a greedy strategy, always optimizing a specific metric.

Transferred to our case, observations correspond to reflected laser rays, clusters correspond to planes, and the metric the algorithm strives to maximize is the measurement probability~\mbox{$p(Z \mid L)$}, which is equivalent to minimizing the error~$E(Z,L)$.
Consequently, in the first step, which assigns each observation its own cluster, we assign each laser reflection its own plane.
As mentioned above, this plane is not parallel to the ray and intersects the ray at its measured length~$r$.
In the following, we call such a plane atomic.
We define the support vector of an atomic plane as the endpoint~\mbox{$s + r v$} of the corresponding ray and the normal vector as the ray direction vector~\mbox{$v$}.
As opposed to atomic planes, regular planes represent not one, but three or more rays.
Therefore, their parameters need to be fitted to the data.

Starting from this trivial maximum likelihood solution, PPE iteratively reduces the number of planes to $J_{\max}$ by merging adjacent planes.
With each merger, the measurement likelihood of the whole scan~\mbox{$p(Z \mid L)$} decreases, whereas the error~$E(Z,L)$ increases by
\begin{align}
	e \coloneqq E(Z,L'') - E(Z,L') \geq 0,
	\label{eq:e}
\end{align}
where $L'$ and $L''$ denote the set of planes before and after the merger.
Greedy as it is, PPE always opts for the merger that incurs the least error increment, which is equivalent to maintaining maximum measurement likelihood.

Due to ambiguities in the decision process, the formulation above will not yield the desired result yet:
The error increment corresponding to merging two or three atomic planes is always zero, because every pair or triple of reflections can be perfectly explained by a single plane.
Therefore, given multiple atomic planes, PPE cannot decide which pair or triple to merge.
Creating a regular plane out of four atomic planes, however, leads to an overdetermined system of equations, hence a regular plane must be fitted to the four reflections, and the corresponding fitting error constitutes the error increment
\begin{align}
	e_{\textup{crt}}(Z,Q) \coloneqq \min_{x,n} E(\{z_q\},\{x,n,Q\}),
	\label{eq:ecrt}
\end{align}
where $Q$ denotes the set of ray indices, and where \mbox{$q \in Q$}.

In order to find the combination of four atomic planes that yields the minimum error increment, PPE needs to assess the fitting errors corresponding to all possible combinations.
For an atomic plane that resides somewhere in the middle of the grid of laser rays, there are \num{17} valid ways to combine it with three of its 4\nobreakdash-connected neighbors, forming so-called tetrominoes: one O-shaped, four T-shaped, four Z-shaped, and eight L-shaped tetrominoes.
The I-shaped tetromino is invalid, because fitting a plane to four endpoints in a straight line again yields ambiguous results.

Once the first regular planes emerge, we can identify two more classes of clustering actions apart from merging tetrominoes: extending a regular plane by an atomic plane, and merging two regular planes.
In each clustering step, PPE must determine the error increment of every possible action, find the one that incurs the least error increment, and merge the respective planes.
The error increment of extending a regular plane by an atomic plane amounts to the difference 
\begin{align*}
	e_{\textup{ext}}(Z,Q,k) 
		\coloneqq e_{\textup{crt}}(Z,Q \cup k) - e_{\textup{crt}}(Z,Q),
\end{align*}
where $Q$ denotes the indices of the rays of the regular plane, and where $k$ is the index of the ray corresponding to the atomic plane.
Merging two regular planes indexed $i$ and $j$ adds 
\begin{align*}
	e_{\textup{mrg}} 
		\coloneqq e_{\textup{crt}}(Z,Q_i \cup Q_j) - e_{\textup{crt}}(Z,Q_i) - e_{\textup{crt}}(Z,Q_j)
\end{align*}
to the total error~$E(Z,L)$.

\subsection{Probabilistic Plane Extraction}

\begin{algorithm}
	\small
	\setstretch{1.1}
	\KwData{$Z$, $J_{\max}$}
	\KwResult{$L$}
	$L \gets \{s_k+r_k v_k, v_k, k\}, k \in \{1,2,\ldots,K\}$ \\
	$(e_{\textup{crt}}, Q_{\textup{crt}}) \gets \mathrm{crt}(Z,L)$ \\
	$e_{\textup{ext}} \gets e_{\textup{mrg}} \gets \infty$ \\
	\While{$J(L) > J_{\max}$}
	{
		\If{$e_{\textup{crt}} = \min(e_{\textup{crt}},e_{\textup{ext}}, e_{\textup{mrg}})$}
		{
			$L \gets L \cup \mathrm{fit}(Z,Q_{\textup{crt}})$ \\
			$L \gets \mathrm{rma}(L, Q_{\textup{crt}})$ \\
			\ElseIf{$e_{\textup{ext}} = \min(e_{\textup{crt}},e_{\textup{ext}},e_{\textup{mrg}})$}
			{
				$L_j \gets \mathrm{fit}(Z,Q_j \cup k)$ \\
				$L \gets \mathrm{rma}(L, \{k\})$ \\
				\Else
				{
					$L_j \gets \mathrm{fit}(Z,Q_i \cup Q_j)$ \\
					$L \gets L \setminus L_i$ \\
				}
			}
		}
		$(e_{\textup{crt}}, Q_{\textup{crt}}) \gets \mathrm{crt}(Z,L)$ \\
		$(e_{\textup{ext}}, j, k) \gets \mathrm{ext}(Z,L)$ \\
		$(e_{\textup{mrg}}, i, j) \gets \mathrm{mrg}(Z,L)$ \\
	}
	\caption{Probabilistic Plane Extraction}
	\label{algo:ppe}
\end{algorithm}

Algorithm~\ref{algo:ppe} provides the PPE pseudocode.
Line~1 initializes the set of atomic planes.
The function~$\mathrm{crt}(Z,L)$ in line~2 loops over all valid tetrominoes of atomic planes and returns the minimum error~$e_{\textup{crt}}$ along with the associated indices~$Q_{\textup{crt}}$.
As there are no regular planes which could be extended or merged at this point, line~3 sets the corresponding error increments~$e_{\textup{ext}}$ and $e_{\textup{mrg}}$ to infinity.
After these initializations, the algorithm starts iteratively reducing the number of planes.
In the first iteration, it always creates a regular plane out of four atomic ones.
This means it first adds the new plane to the map (line~6) and then removes the merged atomic planes (line~7).
Here, the function $\mathrm{fit}(Z,Q)$ fits a plane~$l^*$ to the rays indexed by $Q$:
\begin{align*}
	l^* 
		\coloneqq \mathrm{fit}(Z,Q) 
		\coloneqq \Big\{ \argmin_{x,n} E(\{z_q\}, \{x,n,Q\}), Q \Big\},
\end{align*}
whereas $\mathrm{rma}(L,Q)$ removes the atomic planes indexed by $Q$ from $L$ and returns the updated plane set.
After every manipulation of the plane map, lines~17 to 19 recompute the error increments of all merging options.
To that end, $\mathrm{ext}(Z,L)$ iterates over all possible extensions of all regular planes in $L$ and finds the minimum error increment~$e_{\textup{ext}}$ associated with extending plane~$j$ by ray~$k$.
Similarly, $\mathrm{mrg}$ evaluates for all pairs of neighboring regular planes the hypothetical error increments incurred by merging them and returns the minimum~$e_{\textup{mrg}}$, which corresponds to merging planes~$i$ and $j$.
Lines~9 and 10 update the map during an extension step, while lines~12 and 13 come into play when two regular planes are merged.

For the sake of clarity, algorithm~\ref{algo:ppe} is not optimized.
For an optimized version of PPE, please refer to our MATLAB implementation~\cite{ppe2019}, which features several algorithmic optimizations, optional GPU acceleration, multiple stopping criteria, and a geometric outlier filter.

\section{Experiments}
\label{sec:experiments}

In order to compare PPE with the state of the art, we conduct two series of experiments.
In the first series, we evaluate PPE using the popular SegComp plane extraction benchmark~\cite{hoover1996}.
The deficiencies of this dataset motivated us to create SynPEB, the first publicly available synthetic plane extraction benchmarking dataset, on which we base the second experiment series.

SegComp comprises two collections of organized point clouds, which depict compositions of polyhedral objects on a tabletop.
They were recorded by an ABW structured light sensor and by a Perceptron laser scanner, respectively.
Due to the fact that our measurement model, defined in equation~\eqref{eq:pzL}, is specifically designed for laser sensors, we evaluate our method on the Perceptron collection only.
This dataset is divided into \num{10}~training scans and \num{30}~testing scans.
We use the former to determine the optimum values of $e$ and $d$, the two parameters of the specific PPE version we use in both experiment series.
The parameter $e$, defined in equation~\eqref{eq:e}, denotes the maximum admissible error increment in a clustering step and serves as stopping criterion.
In order to compensate for the high level of noise present in all Perceptron scans, we incorporate a geometric outlier filter in PPE, which prevents clustering neighboring points if their Cartesian distance exceeds a certain threshold~$d$.
To find suitable values for both parameters, we maximize the fraction of correctly segmented planes over a regular grid in $e$ and $d$.

The upper part of table~\ref{tab:evaluation} shows the corresponding experimental results for PPE and compares them to all previous works evaluated on the Perceptron dataset using the performance metrics defined by Hoover et~al.~\cite{hoover1996}.
In order to increase the relevance of the results, we suggest two additional metrics: the $k$-value and the $\RMSE$.
The $k$-value is defined as
\begin{align}
	k \coloneqq \frac{\sum_{j=1}^{J(L)} \hat{K}(l_j)}{K},
	\label{eq:k}
\end{align}
where $\hat{K}(l)$ is a function that takes an extracted plane~$l$ as input, checks if this plane is correctly segmented using the \SI{80}{\percent}~threshold proposed by Hoover et~al., and returns the number of points of the corresponding ground truth plane.
If the input plane is not correctly segmented, the function returns zero.
In this way, $k$ indicates the portion of the point cloud that the algorithm correctly segments into planes.
The root mean squared error~$\RMSE$, defined as
\begin{align}
	\RMSE \coloneqq \sqrt{\frac{E(Z,L)}{J(L)}},
	\label{eq:RMSE}
\end{align}
complements $k$ by providing an estimate of how accurately the extracted planes represent the point cloud.

In addition to quoting the numbers of previous works and stating our results for PPE, we evaluate MSAC and PEAC.
MSAC is a baseline approach based on the RANSAC variant proposed by Torr et~al.~\cite{torr2000}.
Beginning with the input point cloud, this method iteratively detects a plane and removes the inlier points from the cloud until a specified fraction of the original number of points remains.
PEAC -- plane extraction using agglomerative clustering -- refers to the open-source implementation~\cite{peac2018} Feng et~al. provide to complement their paper~\cite{feng2014}.
We are not able to exactly replicate the SegComp results they quote in their paper.
Nevertheless, we state our findings for SegComp in order to establish comparability between our PEAC results across both experiment series.
Analogously to PPE, we determine the optimum parameters for MSAC and PEAC via grid search.
The exact parameter sets for all methods can be found at~\cite{ppe2019}.
Even with these parameters, both MSAC and PEAC return a single false plane detection when processing all testing scans of SegComp, which leads to exploding $\RMSE$-values.
To mitigate this effect, the $\RMSE$-values in table~\ref{tab:evaluation} are based on all planes with \mbox{$\RMSE \leq \SI{10}{m}$} each.

\begin{table*}
	\centering
	\small
	\begin{tabular}{@{}lSSM{2.8em}SSSSS@{}}
		\toprule
		Method & $f~[\si{\percent}]$ & $k~[\si{\percent}]$ & $\RMSE$ $[\si{mm}]$ & $\alpha~[\si{\degree}]$ & $n_o$ & $n_u$ & $n_m$ & $n_s$ \\
		\midrule
		\multicolumn{9}{c}{SegComp Perceptron dataset} \\
		USF~\cite{hoover1996} & 60.9 & \textendash & \textendash & 2.7 & 0.4 & 0.0 & 5.3 & 3.6 \\
		WSU~\cite{hoover1996} & 40.4 & \textendash & \textendash & 3.3 & 0.5 & 0.6 & 6.7 & 4.8 \\
		UB~\cite{hoover1996} & 65.7 & \textendash & \textendash & 3.1 & 0.6 & 0.1 & 4.2 & 2.8 \\
		UE~\cite{hoover1996} & 68.4 & \textendash & \textendash & 2.6 & 0.2 & 0.3 & 3.8 & 2.1 \\
		UFPR~\cite{gotardo2003} & 75.3 & \textendash & \textendash & 2.5 & 0.3 & 0.1 & 3.0 & 2.5 \\
		Oehler et~al.~\cite{oehler2011} & 50.1 & \textendash & \textendash & 5.2 & 0.3 & 0.4 & 6.2 & 3.9 \\
		Holz et.~al.~\cite{holz2013} & 75.3 & \textendash & \textendash & 2.6 & 0.4 & 0.2 & 2.7 & 0.3 \\
		RPL-GMR~\cite{marriott2017} & 72.4 & \textendash & \textendash & 2.5 & 0.3 & 0.3 & 3.0 & 2.0 \\
		Feng et~al.~\cite{feng2014} & 60.9 & \textendash & \textendash & 2.4 & 0.2 & 0.2 & 5.1 & 2.1 \\
		PEAC~\cite{peac2018} & 48.6 & 91.3 & 2.6 & 2.6 & 0.0 & 0.1 & 7.1 & 2.0 \\
		MSAC~\cite{torr2000} & 18.5 & 76.7 & 3.4 & 3.9 & 0.1 & 0.2 & 11.3 & 3.4 \\
		PPE (proposed) & 60.7 & 61.2 & 2.9 & 2.8 & 1.4 & 1.1 & 1.5 & 2.3 \\ 
		\midrule
		\multicolumn{9}{c}{SynPEB dataset} \\
		PEAC~\cite{peac2018} & 29.1 & 60.4 & \tablenum{28.6} & \textendash & 0.7 & 1.0 & 26.7 & 7.4 \\
		MSAC~\cite{torr2000} & 7.3 & 35.6 & \tablenum{34.3} & \textendash & 0.3 & 1.0 & 36.3 & 10.9 \\
		PPE (proposed) & 73.6 & 77.9 & \tablenum{14.5} & \textendash & 1.5 & 1.1 & 7.1 & 16.5 \\ 
		\bottomrule
	\end{tabular}
	\caption{Results of both experiment series.
		The header variables $f$ and $\alpha$ denote the fraction of correctly segmented planes and the mean angular deviation, averaged over all testing scans, while $n_o$, $n_u$, $n_m$, and $n_s$ represent the absolute numbers of oversegmented, undersegmented, missing, and spurious planes compared to the ground-truth segmentation.
		The metrics $k$ and $\RMSE$ are defined in equation~\eqref{eq:k} and \eqref{eq:RMSE}, respectively.
		On average, each scan of the SegComp dataset contains \num{14.6} ground-truth planes, while each scan of the SynPEB dataset is composed of \num{42.6} planes.}
	\label{tab:evaluation}
\end{table*}

\begin{figure}
	\centering
	\begin{subfigure}[t]{\linewidth}
		\includegraphics[width=\linewidth]{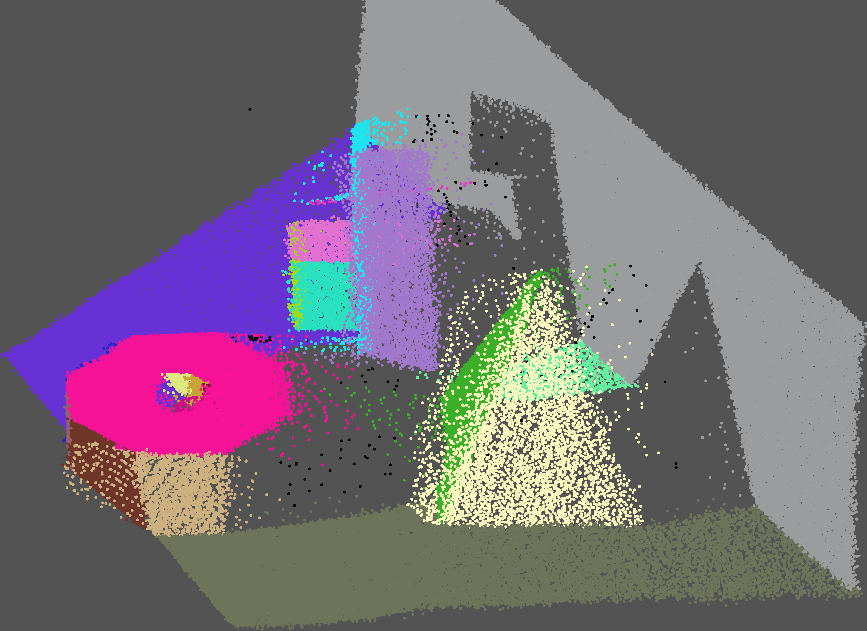}
		\caption{3\nobreakdash-D point cloud colored according to ground-truth segmentation.}
		\label{fig:segcomp_pc}
	\end{subfigure} \\
	\begin{subfigure}[t]{0.49\linewidth}
		\includegraphics[width=\linewidth]{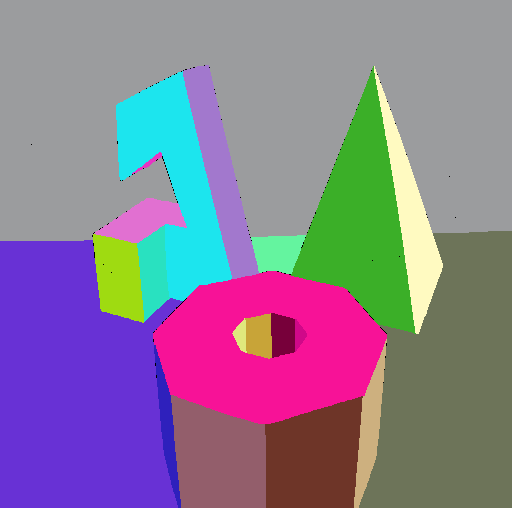}
		\caption{Ground-truth segmentation.}
		\label{fig:segcomp_seg_gt}
	\end{subfigure}
	\hfill
	\begin{subfigure}[t]{0.49\linewidth}
		\includegraphics[width=\linewidth]{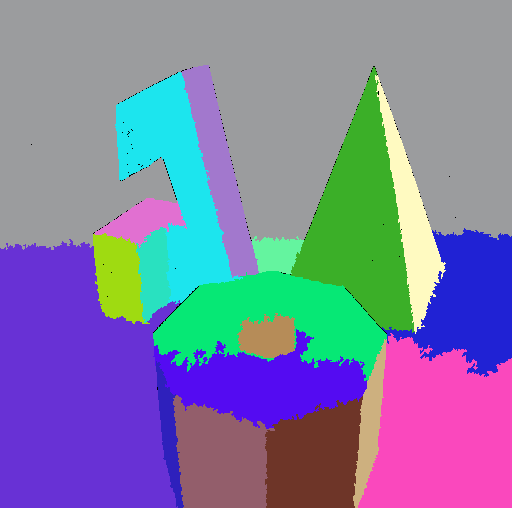}
		\caption{PPE segmentation.}
		\label{fig:segcomp_seg_ppe}
	\end{subfigure}  
	\caption{Point cloud and segmentation images of scan~\texttt{perc.test.23} of the SegComp dataset.
	Outliers are colored black.}
	\label{fig:segcomp}
\end{figure}

Although PPE is designed for maximum accuracy, our method achieves only average results on SegComp.
The reasons lie in the peculiarities of the dataset.
Figure~\ref{fig:segcomp_pc} reveals that the rays that hit an object face at an obtuse angle are much more strongly affected by noise than rays with acute incidence angles, creating the impression that faces with obtuse incidence angles extend in a curved fashion beyond their borders.
Another issue becomes apparent when closely inspecting the ground plane:
Labeling is based on the geometry of the underlying objects, not on the output of the miscalibrated sensor.
The khaki tabletop plane and the purple topside of the octagon in the point cloud in figure~\ref{fig:segcomp_pc}, for example, exhibit kinks due to systematic errors in the lidar calibration.
The labelers, knowing that these planes were flat, labeled both as contiguous planes.
PPE, without knowledge about the real scene, splits each plane into two.
Although desirable, this behavior results in the highest oversegmentation rate among all methods and decreases both the percentage of correctly segmented planes and the $k$-value.

In order to prove that the ground-truth labeling of SegComp is indeed not optimal, we compare the $\RMSE$-values of the ground-truth segmentation to those of PPE.
This time, PPE is configured to extract as many planes from a scan as there are present in the ground truth.
On average, the resulting $\RMSE$-values are \SI{3.2}{\percent} lower than those corresponding to ground truth.
A $t$-test over all scans yields a $p$-value of \SI{12.9}{\%}, which means that the probability of PPE returning a more accurate segmentation than ground truth is as high as \SI{87.1}{\%}.

Similarly to the ground-truth segmentation, the ground-truth angles between adjacent planes were presumably determined based on the underlying data, too:
They are provided as integers rather than as floating-point numbers.

As the aforementioned problems with SegComp bias the evaluation and because there is no publicly available alternative, we created a synthetic plane extraction benchmarking dataset, in short SynPEB, which we use as the basis of the second experiment series.
Like PPE and the implementation of all our experiments, the SynPEB scans and the corresponding sampling engine can be downloaded at \cite{ppe2019}.
The SynPEB world consists of a room of approximately \mbox{$\SI{6}{m} \times \SI{7}{m} \times \SI{3}{m}$} populated with various polyhedral objects, resulting in \num{42.6} planes of different shapes and sizes per scan.
Analogously to SegComp, we divide the dataset into \num{10} training scans and \num{30} testing scans, provided as organized point clouds of \mbox{$500 \times 500$} measurements.
These scans are affected by normally distributed angular noise with standard deviation \mbox{$\sigma_{\textup{ang}} = \SI{1}{mdeg}$} and by normally distributed radial noise with \mbox{$\sigma_{\textup{rad}} = \SI{20}{mm}$}.
Figure~\ref{fig:synpeb} conveys an intuition of what a SynPEB scan looks like.

The lower part of table~\ref{tab:evaluation} shows the plane extraction results for PEAC, MSAC, and PPE on SynPEB.
For the other approaches, there is no working implementation publicly available.
When comparing the results across datasets, we observe that the fraction of planes detected by both PEAC and MSAC is considerably lower on SynPEB than on SegComp.
The high numbers of missed planes indicate that the most likely cause is the challenging nature of SynPEB: 
At almost identical resolutions, SynPEB contains almost three times as many planes per scan as SegComp.
Nevertheless, both the percentage of correctly identified planes and the $k$-value of the PPE results have increased significantly.
The $\RMSE$-value for PPE is \SI{28}{\percent} lower than the radial sensor noise, demonstrating that the method is able to leverage the high number of data points per plane to accurately reconstruct the underlying data.

PPE's high accuracy comes at a price: On average, processing a $500 \times 500$ scan using our open-source implementation takes \SI{1.6}{h} on a single core of an Intel Xeon CPU with \SI{2.6}{GHz}, while our MATLAB implementation of MSAC needs \SI{1.1}{s}.
As a method specifically developed to enable real-time plane extraction, PEAC runs at approximately \SI{30}{Hz} on an Intel i7-7700K processor.

\section{Conclusion and Future Work}
\label{sec:conclusion}

Many authors have investigated the problem of extracting planes from 3\nobreakdash-D laser scans and proposed solutions.
The present paper sets itself apart in two ways.
First, it proposes PPE, an approach to plane extraction that builds upon an accurate probabilistic sensor model instead of the conventional point-to-plane distance heuristic.
Our experiments demonstrate that the accuracy of the sensor model translates to superior plane reconstruction results.
Second, motivated by the deficiencies of the popular plane extraction benchmark SegComp, we suggest an alternative benchmark, dubbed SynPEB.
Both the implementation of the proposed algorithm and the suggested dataset are available online~\cite{ppe2019}.

Due to the promising results, we plan several extensions of PPE.
First of all, we will decrease the runtime to enable online plane extraction.
In addition, we will relax the requirement that the point cloud is organized, and investigate whether leveraging laser remission intensity information can further improve the results.

\bibliographystyle{IEEEtran}
\bibliography{IEEEabrv,plane_extraction.bib}

\end{document}